\documentclass[twocolumn]{article}
\usepackage[a4paper,margin=18mm,footskip=8mm]{geometry}
\usepackage{authblk}
\usepackage{graphicx}
\usepackage{multirow}
\usepackage{amsmath,amssymb,amsfonts}
\usepackage{mathtools}
\usepackage{algorithm, algorithmic}
\usepackage{caption}
\usepackage{subcaption}
\usepackage{amssymb}
\usepackage{stfloats}
\DeclareMathOperator*{\argmax}{arg\,max}
\graphicspath{{images/}}

\usepackage{framed} 
\usepackage{multicol} 
\usepackage{nomencl} 
\makenomenclature
\renewcommand*\nompreamble{\begin{multicols}{2}}
\renewcommand*\nompostamble{\end{multicols}}

\usepackage{etoolbox}
\renewcommand\nomgroup[1]{%
  \item[\bfseries
  \ifstrequal{#1}{A}{Abbreviations}{%
  \ifstrequal{#1}{P}{Environment Parameter Symbols}{%
  \ifstrequal{#1}{S}{Reinforcement Learning Symbols}{}}}%
]}

\usepackage{fancyhdr}
\providecommand{\keywords}[1]
{
  \small	
  \textbf{Keywords---} #1
}

\begin{document}

\title{Data-driven battery operation for energy arbitrage using rainbow deep reinforcement learning
}

\author[1]{Daniel J. B. Harrold\thanks{Email: d.j.b.harrold@keele.ac.uk}}
\author[2]{Jun Cao}
\author[1]{Zhong Fan}
\affil[1]{School of Computing and Mathematics, Keele University, United Kingdom}
\affil[2]{School of Geography, Geology, and the Environment, Keele University, United Kingdom}

\date{\vspace{-5ex}}

\maketitle

\begin{abstract}
As the world seeks to become more sustainable, intelligent solutions are needed to increase the penetration of renewable energy. In this paper, the model-free deep reinforcement learning algorithm Rainbow Deep Q-Networks is used to control a battery in a small microgrid to perform energy arbitrage and more efficiently utilise solar and wind energy sources. The grid operates with its own demand and renewable generation based on a dataset collected at Keele University, as well as using dynamic energy pricing from a real wholesale energy market. Four scenarios are tested including using demand and price forecasting produced with local weather data. The algorithm and its subcomponents are evaluated against two continuous control benchmarks with Rainbow able to outperform all other method. This research shows the importance of using the distributional approach for reinforcement learning when working with complex environments and reward functions, as well as how it can be used to visualise and contextualise the agent's behaviour for real-world applications. \\


\end{abstract}


\begin{table*}[!t]   
	\begin{framed}
		\nomenclature[A]{ANN}{artificial neural network}
		\nomenclature[A]{C51}{categorical deep Q-networks}
		\nomenclature[A]{DDPG}{deep deterministic policy gradients}
		\nomenclature[A]{DQN}{deep Q-networks}
		\nomenclature[A]{DR}{demand response}
		\nomenclature[A]{ESS}{energy storage system}
		\nomenclature[A]{MDP}{Markov decision process}
		\nomenclature[A]{PER}{prioritised experience replay}
		\nomenclature[A]{PV}{photovoltaic}
		\nomenclature[A]{RL}{reinforcement learning}
		\nomenclature[A]{RES}{renewable energy source}
		\nomenclature[A]{TD}{temporal difference}
		\nomenclature[A]{WT}{wind turbine}
		
		\nomenclature[P, 02]{X}{power from demand or generation}
		\nomenclature[P, 01]{$x$}{battery power}
		\nomenclature[P, 03]{$c$}{charge}
		\nomenclature[P, 04]{C}{capacity}
		\nomenclature[P, 05]{P}{energy price}
		\nomenclature[P, 06]{$\eta$}{efficiency}

		\nomenclature[S, 01]{$s$}{state}
		\nomenclature[S, 02]{$a$}{action}
		\nomenclature[S, 03]{$r$}{reward}
		\nomenclature[S, 05]{$\mathcal{S}$}{state-space}
		\nomenclature[S, 06]{$\mathcal{A}$}{action-space}
		\nomenclature[S, 07]{$\mathcal{R}$}{reward function}
		\nomenclature[S, 08]{$\gamma$}{discount factor}
		\nomenclature[S, 09]{$\epsilon$}{random action probability}
		
		\nomenclature[S, 10]{$Q(s,a)$}{action-value}
		\nomenclature[S, 11]{$V(s)$}{state-value}
		\nomenclature[S, 12]{$A(s,a)$}{advantage-value}
		\nomenclature[S, 13]{$\phi$}{shared encoder output}
		\nomenclature[S, 14]{$\theta$}{action-value approximator weights}
		\nomenclature[S, 15]{$\xi$}{state-value approximator weights}
		\nomenclature[S, 16]{$\psi$}{advantage-value approximator weights}
		
		\nomenclature[S, 17]{$p$}{transition priority}
		\nomenclature[S, 18]{$w$}{importance weight}
		
		\nomenclature[S, 19]{$Z(s,a)$}{value distribution}
		\nomenclature[S, 20]{$z$}{support of atom}
		\nomenclature[S, 21]{$d(s,a)$}{distribution probability on atom}
		
		\printnomenclature
	\end{framed}
\end{table*}

\section{Introduction}
As the world seeks to reduce its carbon emissions, increasing the penetration of renewable energy sources (RES) is a necessity to move towards a more sustainable future \cite{bogdanov_low-cost_2021}. However, despite RES being some of the most cost-effective forms of energy generation \cite{ram_comparative_2018}, the intermittent nature of solar photovoltaic (PV) and wind turbine (WT) generation means that non-renewable sources such as natural gas and coal make up the majoirty of our energy networks \cite{department_for_business_energy__industrial_strategy_digest_2020}. Therefore, intelligent solutions are required to manage energy supply and demand more resourcefully.

Smart energy networks combine power engineering with information technologies to manage utilities for improving the efficiency, portability, stability, and security of the grid through the use of demand response (DR) to manipulate energy usage \cite{palensky_demand_2011}. The primary objectives of DR are to increase the utilisation of renewable energy sources (RES) and facilitate energy arbitrage using energy storage systems (ESS), in addition to performing load levelling, peak shaving, and frequency regulation \cite{vazquez_energy_2010}. Although difficult to enable DR at a utility grid scale due to scalability, it is much more feasible to implement in a localised microgrid with its own ESSs and RESs \cite{katiraei_microgrids_2008}. However, this will require an intelligent energy management system to control the ESS able to observe the fluctuating energy demand, intermittent RESs, and volatile wholesale energy pricing markets. 

For this management system, we propose reinforcement learning (RL), a branch of machine learning in which an agent learns to interact with its environment to find an optimal control policy \cite{sutton_reinforcement_2018}. The agent iteratively chooses an action based on observations of the environment to receive a reward with the goal to maximise its total future reward. In addition, artificial neural networks (ANN) can be used to combine RL with deep learning for solutions to more complex problems. The model-free nature of RL algorithms makes them incredibly versatile and easy to implement in situations where there is little information or data available as the agent learns effectively through trial-and-error.

In this paper, we present the use of a state-of-the-art RL algorithm to control an ESS for energy arbitrage in a microgrid. The aim of the agent is to reduce energy costs and consequently increase the utilisation of RES.

\subsection{Related Literature}
Research into applying RL to DR and smart energy networks management has increased significantly over the past decade \cite{vazquez-canteli_reinforcement_2019, perera_applications_2021}.

Before Mnih et al. \cite{mnih_human-level_2015} first introduced modern deep RL, basic tabular methods were the most widely used. Kuznetsova et al. \cite{kuznetsova_reinforcement_2013} used Q-learning \cite{watkins_learning_1989} to control an ESS connected to WT generation to increase its utilisation under a number of scenarios. However, an issue with tabular RL is that the observation values must be from a discrete state-space. Kofinas et al. \cite{kofinas_fuzzy_2018} use fuzzy logic to adapt Q-learning to a continuous state and action space, but more modern deep RL algorithms can fulfil this too.

Authors generally have moved away from tabular methods to deep RL methods using ANNs as functional approximators, such as Deep Q-Networks (DQN) \cite{mnih_human-level_2015}. Examples of using DQN for energy arbitrage and energy cost reduction include Fran\c{c}ois-Lavet et al. \cite{francois-lavet_deep_2016} with a hybrid ESS setup in a solar microgrid, Bui et al. \cite{bui_double_2020} with a variant of DQN considering uncertainities in the environment, and Cao et al. \cite{cao_deep_2020} with another more advanced variant based on a realistic lithium-ion battery degradation model. Although value-function methods such as Q-learning and DQN are simple algorithms with robust learning properties, a limitation is that the agent must select from a discrete set of actions which could be considered unsuitable for continuous control problems; such as ESS operation.

As a result, many authors have instead investigated actor-critic methods. These are a branch of RL where one ANN is used to choose a continuous value for the action and a second ANN is used to evaluate performance.
Examples include Mocanu et al. \cite{mocanu_-line_2018} using deep deterministic policy gradient (DDPG) \cite{lillicrap_continuous_2016} to schedule a building's energy management system,
Zhang et al. \cite{zhang_data-driven_2021} used proximal policy optimisation (PPO) \cite{schulman_proximal_2017} for energy cost reduction using ESS control in a AC/DC hybrid microgrid, and Pinto et al. \cite{pinto_coordinated_2021} used soft actor-critic \cite{haarnoja_off-policy_2018} for cost reduction and load levelling across multiple smart buildings. However, actor-critic methods are significantly more sensitive to hyperparameter tuning than value-function methods and can suffer from unstable and unpredictable learning \cite{sutton_reinforcement_2018}. Therefore, the correct balance must be found between control resolution and robust learning.

\subsection{Contributions}
The contributions of this work are as follows:

\begin{itemize}
\item The paper presents the state-of-the-art deep RL algorithm Rainbow DQN \cite{hessel_rainbow:_2018} for the control agent. To the best of our knowledge, this is the first time the algorithm has been used for a DR application and is tested against two continuous control benchmarks: an actor-critic method and a linear programming model.

\item The paper assesses the distributional approach of deep RL where the agent evaluates performance using a value distribution, presenting the benefits of using this approach for DR applications.

\item The methods are evaluated under four scenarios with different sizes of state and action-spaces to evaluate how different algorithms utilise forecasting as well as having a greater resolution of control.

\item The demand and weather data used in the simulations were collected from the same site at Keele University Campus as part of the Smart Energy Network Demonstrator project \cite{keele_university_smart_2020}, as well as using dynamic energy prices from a wholesale energy market \cite{nord_pool_historical_2020}.
\end{itemize}

\subsection{Structure}
The rest of this paper is organised into a description of the microgrid environment in Section \ref{sec:SEN}, the background behind the RL methodology in Section \ref{sec:RL}, outlining the case study in Section \ref{sec:methods}, results and discussions in Section \ref{sec:results} with final conclusions in Section \ref{sec:conclusion}.
\section{Microgrid Description}
\label{sec:SEN}

\begin{figure*}[!t] 
	\centering
	\includegraphics[width=.90\textwidth]{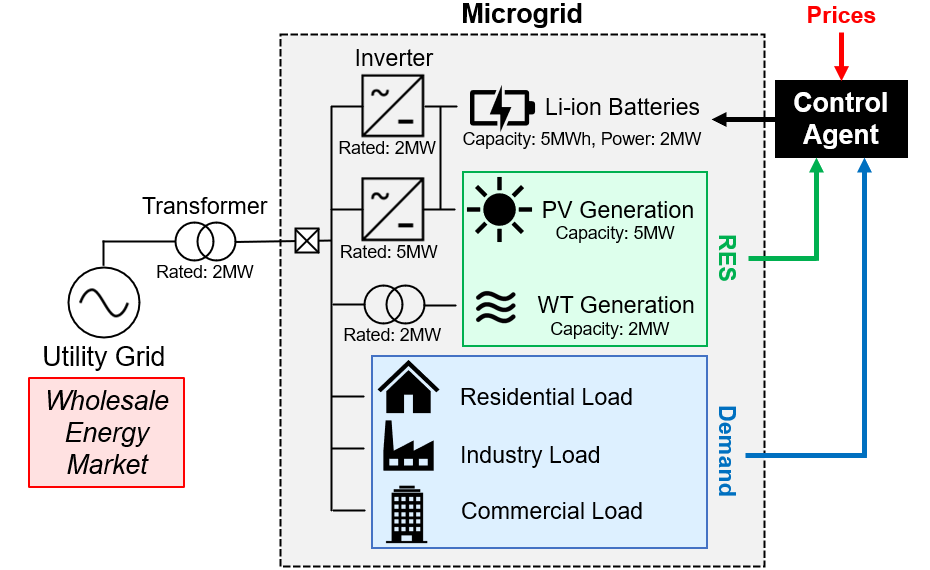}
	\caption{Basic schematic of the microgrid.} \label{fig:grid}
\end{figure*}

Microgrids are localised energy networks with their own demand and RES that are able to trade energy with the main utility grid or operate as an independent island \cite{katiraei_microgrids_2008}. This section will describe the microgrid used in the future case studies and how it is modelled as a Markov decision process (MDP) for RL.

\subsection{Environment Description}
The microgrid is fitted with a large lithium-ion battery ESS as well as both PV and WT generation. It is connected to the main utility grid that sets dynamic energy prices from which the microgrid can import energy from or export to. AC and DC power lines are present connected via inverters, as well as transformers between the microgrid and both the utility grid and the WT. The basic schematic of the environment is shown in Figure \ref{fig:grid}.

\subsubsection{Demand and Pricing}
The demand data collected at Keele University Campus ranges from 00:00 January 1st 2014 up to and including 23:00 December 31st 2017. The raw data is separated into different residential, industrial, and commercial sites as well as readings for key individual buildings. The demand used for this simulation is from the three main incomer substations into the campus with the half-hourly readings summed to match the frequency of the hourly weather data. All loads are AC.

The microgrid operates under a dynamic energy pricing scheme. The energy prices from the utility grid are from a real day-ahead energy trading market covering the UK \cite{nord_pool_historical_2020} but also with a set maximum price of $250\text{GBP/MWh}$. The price for exporting to the grid is the same as the import price.

\subsubsection{Renewable Energy Sources}
The output for the renewable generation is modelled using weather data collected at the Keele University weather station. The WT's output $\text{X}^{\text{WT}}$ uses power curve modelling with wind speed $v$:

\begin{equation}
\text{X}^{\text{WT}} = 
\begin{cases}
	0 & v < v_{\text{ci}} \text{ or } v > v_{\text{co}}\\
	\frac{1}{2}\rho \pi r^2 c_{p}v^{3} & v_{\text{ci}} \leqslant v < v_{r} \\
	\text{X}_{r}^{\text{WT}} & v_{r} \leqslant v \leqslant v_{\text{co}}
\end{cases}
\end{equation}

where $v_{ci}=3\text{ms}^{-1}$, $v_{r}=12\text{ms}^{-1}$, and $v_{co}=25\text{ms}^{-1}$ are the cut-in, rated, and cut-out wind-speeds respectively. These values are comparable to a 1MW turbine with a blade radius of $r=30\text{m}$ and a power coefficient $c_p=0.4$ \cite{carrillo_review_2013}. This microgrid considers two of these turbines for a combined maximum wind capacity of $2\text{MW}$. The same power curve approach is used by both Kuznetsova et al. \cite{kuznetsova_reinforcement_2013} and Zhang et al. \cite{zhang_data-driven_2021} for their RES output. The solar output is calculated using hourly solar radiation data and scaled to model a solar farm with a maximum capacity of 5MW. The output of the WT is AC while the output of the PV is DC.

\subsubsection{ESS}
\begin{figure*}[!t]
\centering
	\begin{subfigure}{.32\textwidth}
		\centering
		\includegraphics[width=\textwidth]{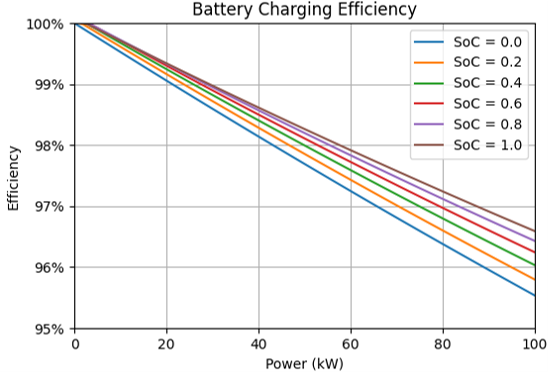}
		\caption{Battery efficiency when charging.}
		\label{fig:essch}
	\end{subfigure}
	\begin{subfigure}{.32\textwidth}
		\centering
		\includegraphics[width=\textwidth]{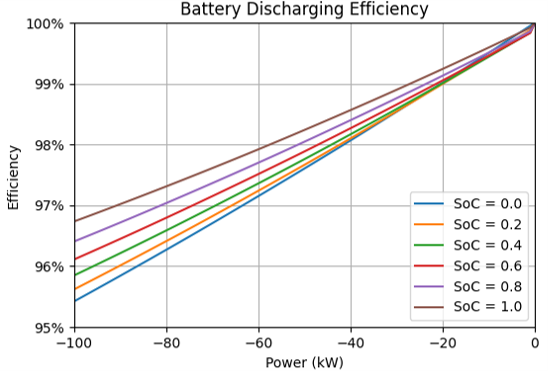}
		\caption{Battery efficiency when discharging.}
		\label{fig:essdis}
	\end{subfigure}
	\begin{subfigure}{.32\textwidth}
		\centering
		\includegraphics[width=\textwidth]{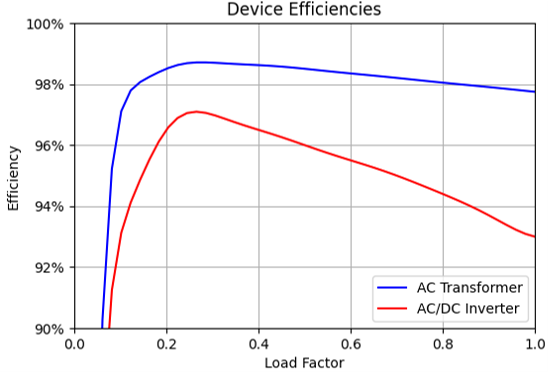}
		\caption{Transformer and inverter efficiencies.}
		\label{fig:invtra}
	\end{subfigure}
	\caption{Efficiency profiles of different components of the microgrid.}
\end{figure*}

The charge of the ESS is bound between 0 and its maximum capacity $\text{C}_{\text{max}}=5\text{MWh}$, with the amount of energy the battery can charge or discharge each step bound between 0 and its maximum power $\text{X}_{\text{max}}=2\text{MW}$. The charge of the ESS $c_t$ is calculated using:

\begin{equation}
c_t = (x_t\eta^{\text{ESS}}_t + c_{t-1}) \eta^{\text{SDC}}_t
\end{equation}

The ESS's charging and discharging efficiencies are nonlinear and modelled as a function of the current charge and operating power using the same approximation as Morstyn et al. \cite{morstyn_model_2018} with the charging and discharging profiles shown in Figures \ref{fig:essch} and \ref{fig:essdis} respectively. For this, the 5MWh ESS is modelled as 50 100kWh lithium-ion batteries which charge and discharge in unison. The ESS also has a linear self-discharge efficiency of $\eta_{\text{SDC}}=0.999$ such that it loses $0.1\%$ of its charge every hour. The output of the ESS is DC.

\subsubsection{Transformer and Inverter}
The case study microgrid is a hybrid AC/DC network with inverters between the two power lines and transformers to step-down the voltage to the microgrid. The efficiency profiles are nonlinear and given as a function of the load factor; the current load divided by the rated load. The rated loads of each of the transformers and inverters can be found in Figure \ref{fig:grid} with the efficiency profiles modelled using a polynomial estimation, shown in Figure \ref{fig:invtra}. Although the real efficiency of these devices at low load factor would approach 0\%, the minimum efficiency is capped at 10\% to prevent divisions by 0 during environment calculations. 

Note that the ESS and PV generation are connected to both 2MW and 5MW rated inverters as well as to each other. When energy is to pass between the AC and DC lines, the control agent will automatically use whichever inverter that would result in lower power losses. Mbuwir et al. \cite{mbuwir_reinforcement_2019} used a similar profile for inverters \cite{driesse_beyond_2008} for RL in a microgrid with solar energy.

\subsection{Markov Decision Process}
For RL to operate the microgrid environment, the control process must be modelled as an MDP. This means the control process must satisfy the Markov property and can be generalised to a a state-space $\mathcal{S}$, action-space $\mathcal{A}$, reward function $\mathcal{R}$, and a state transition function. The rest of this section will explain the states, actions, and rewards for this case study.

\subsubsection{States}
The simulations in Section \ref{sec:results} use agents with two different state-spaces. The basic agents makes a total of 8 observations: the ESS charge $c_t$, grid demand $\text{X}^{\text{D}}$, energy price $\text{P}$, PV generation $\text{X}^{\text{PV}}$, WT generation $\text{X}^{\text{WT}}$, as well as the hour in the day, week, and if it was a workday or not. The forecasting agents also observe forecasts calculated using ANNs based on previous values and weather data. This means the advanced agents observe the same 8 observations as well as predictions for the next hour of demand, solar and wind generation, and price for 12 total observations. As the agent does not observe every aspect of the underlying state, the process is treated as a partially observable MDP.

Each of the values are scaled and bound between 0 and 1 to improve the training stability of the ANNs. The charge, demand, PV and WT generation are all divided by the maximum capacity of the battery 5\text{MWh}. The dynamic energy price is divided by the set price limit of 250GBP/MWh. The hourly values begin at 0 and end at 1 for the day or week. The workday value is set to 0 for weekends or UK bank holidays; otherwise 1.

\subsubsection{Actions}
The action the agent selects controls the ESS's power $x$ at that step. The agent can select from several discrete actions where the ESS can charge ($x>0$), discharge ($x<0$), or can remain idle ($x=0$). Two different actions-spaces of different size are considered: the first with 5 actions and the second with 9 actions. The actions are evenly spaced between the maximum charging and discharging power such that the action-space of size 5 is:
\begin{equation}
\mathcal{A} = [2\text{MW}, 1\text{MW}, 0\text{MW}, -1\text{MW}, -2\text{MW}]
\end{equation}

Although a real ESS would be able to charge or discharge by any continuous value from its minimum to its maximum power, restricting the size of the action-space improves learning and performance in value-function RL methods as the agent has to explore fewer state-action pairs \cite{sutton_reinforcement_2018}.

Due to the configuration of the grid shown in Figure \ref{fig:grid} the ESS will always charge from RES first before importing from the utility grid to minimises energy losses from the transformer and inverter.

\subsubsection{Reward}
The reward at each step is equal to the energy cost savings made by the agent. As the ESS will always look to charge from the RES generation first, the excess PV generation $\text{X}^{\text{PV+}}$ and excess WT generation $\text{X}^{\text{WT+}}$ represent the amount of RES left after the charging has been considered. From this, the demand of the DC portion of the grid $\text{X}^{\text{dc}}$ is calculated as:

\begin{equation}
\text{X}^{\text{dc}} = (x\eta^{\text{ESS}} - \text{X}^{\text{PV+}})\eta^{\text{inv}}_{\text{[2,5]MW}}
\end{equation}

This is then used to calculate the total amount of energy being imported from the utility grid $\text{X}^{\text{in}}$:

\begin{equation}
\text{X}^{\text{in}} = (\text{X}^{\text{dc}} + \text{X}^{\text{D}} - \text{X}^{\text{WT}}\eta^{\text{tra}}_{\text{2MW}})\eta^{\text{tra}}_{\text{2MW}}
\end{equation}

During preliminary studies it was found that the agent would fail to learn an optimal policy properly from this value alone because the agent would be rewarded if the demand was low with high renewable generation and punished in the reverse case. This is an issue because these are environmental factors which the agent has no control over so leads to problems with credit assignment. Therefore, a base demand $\text{X}^{\text{base}}$ is used to normalise the reward:

\begin{equation}
\text{X}^{\text{base}} =  (\text{X}^{\text{D}} - \text{X}^{\text{PV}}\eta^{\text{inv}}_{\text{[2,5]MW}} - \text{X}^{\text{WT}}\eta^{\text{tra}}_{\text{2MW}})\eta^{\text{tra}}_{\text{2MW}}
\end{equation}

The agent should also be discouraged from taking an action that would exceed the ESS's charge boundaries. In this event, the agent receives a negative reward $R_{\text{punish}}$ equal to the squared amount it would have exceeded the boundaries divided by the maximum power $\text{X}_{\text{max}}$. This gives a combined reward function:

\begin{equation}
\mathcal{R} = \frac{\text{P}_t(\text{X}^{\text{base}}_t - \text{X}^{\text{in}}_t)}{\text{P}_{\text{max}}\text{C}_{\text{max}}} - R_{\text{punish}}
\end{equation}

The energy savings value is divided by the product of the maximum price $\text{P}_{\text{max}}$ and the ESS capacity $\text{C}_{\text{max}}$ so that the reward largely remains between -1 and 1 for stable learning. This reward function is different to using pure energy arbitrage in that the nonlinear efficiencies mean the demand and RES are taken into account as well. If the efficiencies were constant then the reward function would cancel them out and the agent would be performing straight arbitrage. However, in this case the agent considers both arbitrage as well as how to optimally utilise RES taking into account the battery, inverter, and transformer efficiencies. 
\section{Reinforcement Learning Methodology}
\label{sec:RL}
This section will explore the theory behind RL and introduce the specific algorithms used in the simulations later in the paper.

The key benefit of RL is that the methods are tpically model-free meaning the state transition function and reward function do not need to be known. This makes them incredibly versatile and easy to implement as the agents learn entirely from its own experience; effectively through trial-and error \cite{sutton_reinforcement_2018}. Also, model-based solutions are only as accurate as the model they are based on. 

\subsection{Fundamentals}
\begin{figure}[!t] 
	\centering
	\includegraphics[width=.45\textwidth]{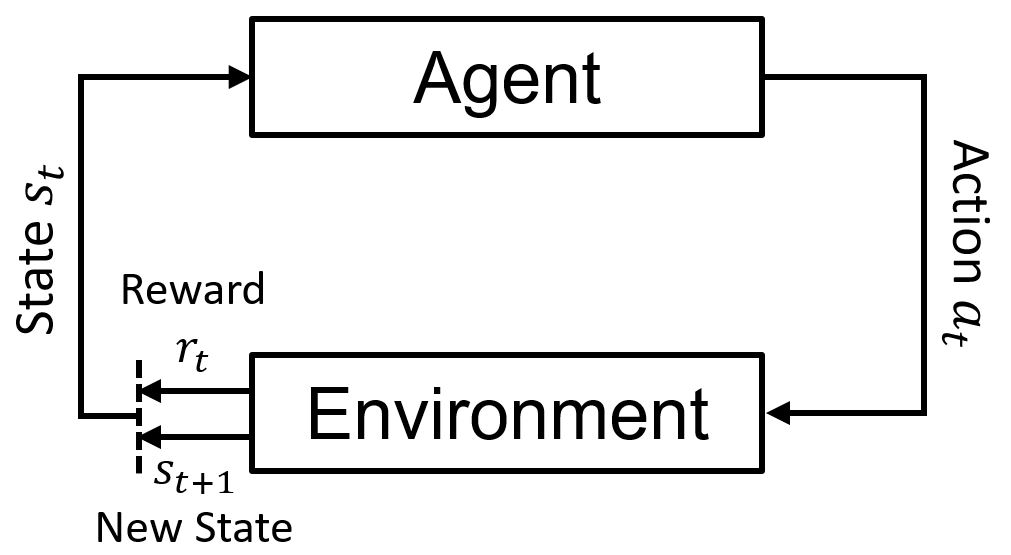}
	\caption{Reinforcement learning control process.}
	\label{fig:rl}
\end{figure}

Each time-step $t$ in RL, the agent observes the current state of the environment $s_t$ from its state-space $\mathcal{S}$ and selects an action $a_t$ from an action-space $\mathcal{A}$ following a learnt policy. The agent then receives a reward $r_t$ from a reward function $\mathcal{R}$ and transitions to a new state $s_{t+1}$ following the state transition function. This is shown in Figure \ref{fig:rl}. The goal of the agent is to learn an optimal policy that maximises its total discounted future reward, scaled with a discount factor $\gamma$.

\begin{equation}
R_t = r_{t}+\gamma r_{t+1}+\gamma^2 r_{t+2}+\ldots = \sum\limits_{k=0}^{\infty}\gamma^{k}r_{t+k}
\end{equation}

Value-function methods evaluate performance by assigning value to each state or state-action pair. The state-value $V(s)$ represents how valuable it is to be in a given state and the action-value $Q(s,a)$ how valuable it is to take a particular action in that state. The control problem is solved recursively using the Bellman equation \cite{bellman_theory_1954}:

\begin{equation}
Q(s_t, a_t) = \mathbb{E} \mathcal{R}(s_t, a_t) + \gamma \mathbb{E}Q(s_{t+1},a_{t+1})
\end{equation}

In these methods, the agent will follow a deterministic policy where the agent will always choose the action it predicts has the highest action-value. For exploration, value-function methods typically follow an $\epsilon$-greedy policy where each time-step the agent has a probability of $\epsilon$ to choose a random action and $(1-\epsilon)$ to select the action with the highest action-value. The $\epsilon$ value will typically decrease over time where maximising the return becomes more important than exploration.

\subsection{Algorithms}
\label{section:alg}
This section will outline the methods used in the simulation. Each one is a valuation function method using an ANN as a functional approximator so are able to process continuous observation values, rather than tabular RL methods which require discrete state and action-spaces.

\subsubsection{Deep Q-Networks (DQN)}
Q-learning \cite{watkins_learning_1989} is a tabular temporal difference (TD) learning method and had been the most widely used RL algorithm prior to modern deep RL methods. DQN \cite{mnih_human-level_2015} is an adaptation of Q-learning using a functional approximator to predict the action-values and then updating the network weights $\theta$ using stochastic gradient descent. TD methods such as these learn using the TD-error $\delta$ calculated as the difference between the current state-value and a target value:

\begin{equation}
\delta_i \doteq r_{i}+\gamma Q(s_{i+1},a_{i+1}) -Q(s_i,a_i)
\end{equation}

It is common practice to use a stable target network to calculate the target action-value $\hat{Q}(s,a)$ with network weights $\hat{\theta}$ frozen from the evaluation network at a previous step and then update $\hat{\theta}\leftarrow \theta$ every $C$ steps. A target value $y_i$ for the Bellman update is calculated by:

\begin{equation}
y_i = r_i + \gamma \max_{a \in \mathcal{A}}\hat{Q}_{\hat{\theta}}(s_{i+1},a)
\end{equation}

Experience replay \cite{lin_self-improving_1992} is also used to improve learning through the use of a replay buffer. Each step, the transition $\langle s_t,a_t,r_t,s_{t+1} \rangle$ is stored in the buffer and then the evaluation network is trained by sampling a random batch of those tuples. The evaluation network is updated using the mean squared error loss which corresponds to the TD-error $\delta$:

\begin{equation}
\mathcal{L}_i(\theta) = \mathbb{E} \left[(y_i - Q_{\theta}(s_i,a_i))^2\right]
\end{equation}

Many environments will reach a terminal state and then reset for the next episode. In this microgrid environment, there are no terminal states so the algorithm continues until finished. The pseudocode for DQN is shown in Algorithm \ref{alg:dqn}.
\begin{algorithm}[t]
\caption{Deep Q-Networks}
\begin{algorithmic}[l]
\STATE Initialise network $Q_{\theta}(s,a)$ with random weights $\theta$
\STATE Initialise target network $\hat{Q}_{\hat{\theta}}(s,a)$ and set $\hat{\theta}\leftarrow \theta$
\STATE Initialise replay buffer

\FOR{$t=0, T$}
	\STATE Observe state $s_t$
	\STATE \textbf{if} probability $p<\epsilon$, choose random action $a_t$
	\STATE \textbf{else} $a_t = \argmax_a Q_{\theta}(s_{t},a_t)$
	\STATE Execute $a_t$, observe reward $r_t$ and next state $s_{t+1}$
	\STATE Store transition $\langle s_t,a_t,r_t,s_{t+1} \rangle$
	\STATE Sample batch $\langle s_i,a_i,r_i,s_{i+1} \rangle$ randomly
	\STATE Calculate targets $y_i = r_i+\gamma \max_a\hat{Q}_{\hat{\theta}}(s_{i+1},a_i)$
	\STATE Loss $\mathcal{L}_i(\theta) = \mathbb{E} \left[(y_i - Q_{\theta}(s_i,a_i))^2\right]$
	\STATE Perform gradient descent on $\theta$ using loss
	\STATE \textbf{if} $\epsilon > \epsilon_{min}$, $\epsilon \leftarrow (1-\epsilon_{decay})\epsilon$
	\STATE Every $C$ steps, set $\theta^{-}\leftarrow \theta$
\ENDFOR
\end{algorithmic} \label{alg:dqn}
\end{algorithm}

\subsubsection{Double Deep Q-Networks (DDQN)}
An issue with standard DQN is that the target network selects the next-best action with the maximisation step which is then itself used for evalualtion. This can lead to overestimations in the value calculations causing the agent to prefer those particular actions. DDQN \cite{van_hasselt_deep_2016} decouples this by removing the maximisation step and instead using the evaluation network to predict the next-best action:

\begin{equation}
a^{*} = \argmax_{a\in\mathcal{A}}Q_{\theta}(s_{i+1},a)
\end{equation}

The benefit of this is that the agent is not automatically assumed to always take the most valuable action from the next state. The updated target is then calculated by:

\begin{equation}
y_i = r_i + \gamma \hat{Q}_{\hat{\theta}}(s_{i+1},a^{*})
\end{equation}

Bui et al. \cite{bui_double_2020} use DDQN to manage a community battery ESS in a microgrid, showing how the algorithm can perform well despite the uncertainties in microgrid environments.

\subsubsection{Double Dueling Deep Q-Networks (D3QN)}
Using only $Q(s,a)$ for the Bellman update does not take into account the inherit value of the state $V(s)$. Therefore, the advantage function $A(s,a)$ separates the two to give greater merit for taking more valuable actions when in low value states. Dueling DQN \cite{wang_dueling_2016} builds on DQN by using dueling ANN architecture using a shared encoder with two output streams: one to estimate $A(s,a)$ for each of the actions with weights $\psi$ and the other to estimate $V(s)$ with weights $\xi$. From this, $Q(s,a)$ can be calculated with:

\begin{equation}
Q_{\theta}(s_i,a_i) = V_{\xi}(\phi_i)+A_{\psi}(\phi_i,a_i)-\bar{A}_{\psi}(\phi_{i})
\end{equation}

using the mean advantage value across all actions $\bar{A}(\phi_{i})$ for identifiability. Here, $\phi$ is the output of the shared encoder which are the hidden layers of the functional approximator that both output streams share:

\begin{equation}
\phi_i = f_{\zeta}(s_i)
\end{equation}

Therefore, the weights of $Q(s,a)$ are $\theta=\lbrace \zeta, \xi, \psi \rbrace$ in their concatenation.

In previous work \cite{harrold_battery_2020}, we used D3QN to control a battery in a simple microgrid setup with RES under a dynamic wholesale energy price where it outperformed both DQN and DDQN. 

\subsubsection{Prioritised Experience Replay DQN (PER-DQN)}
Random sampling of the replay buffer can be inefficient as newer transitions will not be sampled as frequently as older transitions. A prioritised experience replay (PER) \cite{schaul_prioritized_2016} buffer samples transitions by assigning a priority $p$ to each transition such that more important transitions are sampled quicker and more frequently. Each priority is equal to the absolute TD-error $\delta$ of its most recent sample plus a small offset with a probability $P$ of that transition being sampled:
\begin{equation}
P(i) = \frac{p_i^\alpha}{\sum_k p_k^\alpha}
\end{equation}

with a priority scale $\alpha$. An increasing importance scale $\beta$ is then used to calculate the importance weight $w$ which scales the loss in training to give more attention to transitions with larger errors:
\begin{equation}
w_i = \left(\frac{1}{N} \times \frac{1}{P(i)}\right)^\beta / \max_j w_j
\end{equation}

Yang et al. \cite{yang_deep_2020} used D3QN with PER to control an ESS in a wind farm to account for uncertainties in the wind power generation and wholesale energy price.

\subsubsection{Multistep DQN (MS-DQN)}
In some environments, taking the return over multiple steps for the target calculation can increase learning speed \cite{sutton_learning_1988} as the agent follows the current learnt policy for $n$-steps and then performs the backup over that time-period. The value of $n$ should be tuned for different environments. The new target $y_i$ is then calculated using:

\begin{equation}
y_i = R_{t:t+n-1} + \gamma^n \max_{a\in\mathcal{A}} \hat{Q}\left(s_{t+n}, a; \theta^-\right)
\end{equation}

Kuznetsova et al. \cite{kuznetsova_reinforcement_2013} used a 3-step return for tabular Q-learning for 2-step-ahead battery scheduling with scenario-based wind power generation. However, to our knowledge, there is no case of MS-DQN being used for DR.

\subsubsection{NoisyNet DQN (NN-DQN)}
This uses a pure-greedy policy but replaces the linear layers of the DQN ANN with noisy layers that inject noise into the $Q(s,a)$ calculation. The benefit of this is to an $\epsilon$-greedy policy is that the functional approximator can learn to ignore the noise over time in different parts of the state-space rather than having a policy-wide exploration rule. Using Gaussian noise parameters $\epsilon^w$ and $\epsilon^b$ and an initial sigma value $\sigma_0$, the output of the noisy layer is calculated as:

\begin{equation}
y =\left(\mu^w + \sigma^w \odot \epsilon^w\right)x + \mu^b + \sigma^b \odot \epsilon^b
\end{equation}

Cao et al. \cite{cao_deep_2020} used NN-DDQN for energy arbitrage using a realistic lithium-ion battery model and found faster learning properties than both DQN and D3QN.

\subsubsection{Categorical DQN (C51)}
An issue with using the expectation value for $Q(s,a)$ is that it can over-generalise in environments with a non-deterministic reward function \cite{bellemare_distributional_2017}. Categorical DQN removes the expectation step in the Bellman equation and instead estimates the value distribution $Z(s,a)$:

\begin{equation}
Z(s_t, a_t) \stackrel{D}{=} \mathcal{R}(s_t, a_t) + \gamma Z(s_{t+1},a_{t+1})
\end{equation}

A number of atoms are assigned to each action for $N_{\text{actions}} \times N_{\text{atoms}}$ output neurons with the probability distribution of the return $d(s,a)$ calculated using a softmax applied separately across the atoms for each action. The algorithm C51 \cite{bellemare_distributional_2017} uses 51 atoms across each action. The probability mass on each atom is equal to:

\begin{equation}
d_i(s,a) = \frac{\exp(\theta_i(s,a))}{\sum_j \exp(\theta_j(s,a))}
\end{equation}

Each of these probability masses can then be projected onto a support $z$ with a value equally spaced between a minimum return $v_{\text{min}}$ and a maximum return $v_{\text{max}}$:

\begin{equation}
z_i = v_{\text{min}} + (i-1)\frac{v_{\text{max}}-v_{\text{min}}}{N_{\text{atoms}}-1}
\end{equation}

The sum of these probability masses on each support is equal to $Q(s,a)$. The target used for training is the projected target distribution calculated using the categorical algorithm \cite{bellemare_distributional_2017}:

\begin{equation}
(\Phi \hat{\mathcal{T}} Z_{\hat{\theta}}(s,a))_i = \sum^{N_{\text{atoms}}}_{j=0} \left[ 1 - \frac{\vert [\hat{\mathcal{T}}z_j]^{v_{\text{max}}}_{v_{\text{min}}}-z_i \vert}{\Delta z} \right]^1_0 \hat{d}_j(s_{i+1},a^{*})  
\end{equation}

where $\Phi$ is the projection of the distribution onto $z$ and $\mathcal{T}$ is the distributional Bellman operator. As this generates a discrete probability distribution, the Kullback-Leibler divergence between the predicted value and the projected target is used for the loss function:

\begin{equation}
\mathcal{L}_i(\theta)=D_{\text{KL}}\left( \Phi \hat{\mathcal{T}} Z_{\hat{\theta}}(s_i,a_i) \Vert Z_{\theta}(s_i,a_i)\right)
\end{equation}

To the best of our knowledge, no authors have used C51 or a similar distributional RL algorithm for DR applications.

\subsubsection{Rainbow DQN}
Rainbow \cite{hessel_rainbow:_2018} combines all of the components just described into a single algorithm. The noisy linear layer has $N_{\text{actions}} \times N_{\text{atoms}}$ output neurons for the advantage stream and $N_{\text{atoms}}$ for the state-value stream to implement the dueling architecture into the value distribution calculation. The new distribution probability on an atom is then calculated by:

\begin{equation}
d_i(s,a) = \frac{\exp(\xi_i(\phi) + \psi_i(\phi,a) - \bar{\psi}_i(\phi))} {\sum_j\exp(\xi_j(\phi) + \psi_j(\phi,a) - \bar{\psi}_j(\phi))}
\end{equation}

The multistep return is used with the next-best action to calculate the target distribution used for the Kullback-Leibler divergence loss function. The transitions are stored in a PER buffer but instead uses the divergence error to update the priority with a smaller offset.

As with C51, we cannot find an example of authors using the complete form of Rainbow DQN for DR. The pseudocode for Rainbow is shown in Algorithm \ref{alg:rainbow}.
\begin{algorithm}[t]
\caption{Rainbow Deep Q-Networks}
\begin{algorithmic}[l]

\STATE Initialise network $Z_{\theta}(s,a)$ with random weights $\theta$
\STATE Initialise target network $Z_{\hat{\theta}}(s,a)$ and set $\hat{\theta}\leftarrow \theta$
\STATE Initialise PER buffer with initial priority $p_0=1$

\FOR{$t=0, T$}
	\STATE Observe state $s_t$
	\STATE Choose action $a_t = \argmax_a \sum_i z_i d_i (s_t, a)$
	\STATE Execute $a_t$, observe reward $r_t$ and next state $s_{t+1}$
	\STATE Store transition $\langle s_t,a_t,r_t,s_{t+1} \rangle$ with $p_t=\max_{i<t}p_i$
	\STATE Sample batch $\langle s_i,a_i,r_i,s_{i+1} \rangle$ with $P(i)=p_i^\alpha / \sum_j p_j^\alpha$
	\STATE Calculate weights $w_i = (N \cdot P(i))^{-\beta} / \max_j w_j $
	\STATE Calculate projected targets $\Phi \hat{\mathcal{T}} Z_{\hat{\theta}}(s_i,a_i)$
	\STATE Loss $\mathcal{L}_i(\theta)=w_i \cdot D_{\text{KL}}(\Phi \hat{\mathcal{T}} Z_{\hat{\theta}}(s_i,a_i) \Vert Z_{\theta}(s_i,a_i))$
	\STATE Perform gradient descent on $\theta$ using loss
	\STATE Update $p_i \leftarrow D_{\text{KL}}(\Phi \hat{\mathcal{T}} Z_{\hat{\theta}}(s_i,a_i) \Vert Z_{\theta}(s_i,a_i))$
	\STATE Every $C$ steps, set $\hat{\theta}\leftarrow \theta$
\ENDFOR
\end{algorithmic} \label{alg:rainbow}
\end{algorithm}
\section{Case Study}
\label{sec:methods}

\begin{figure*}[!t] 
	\centering
	\includegraphics[width=.95\textwidth]{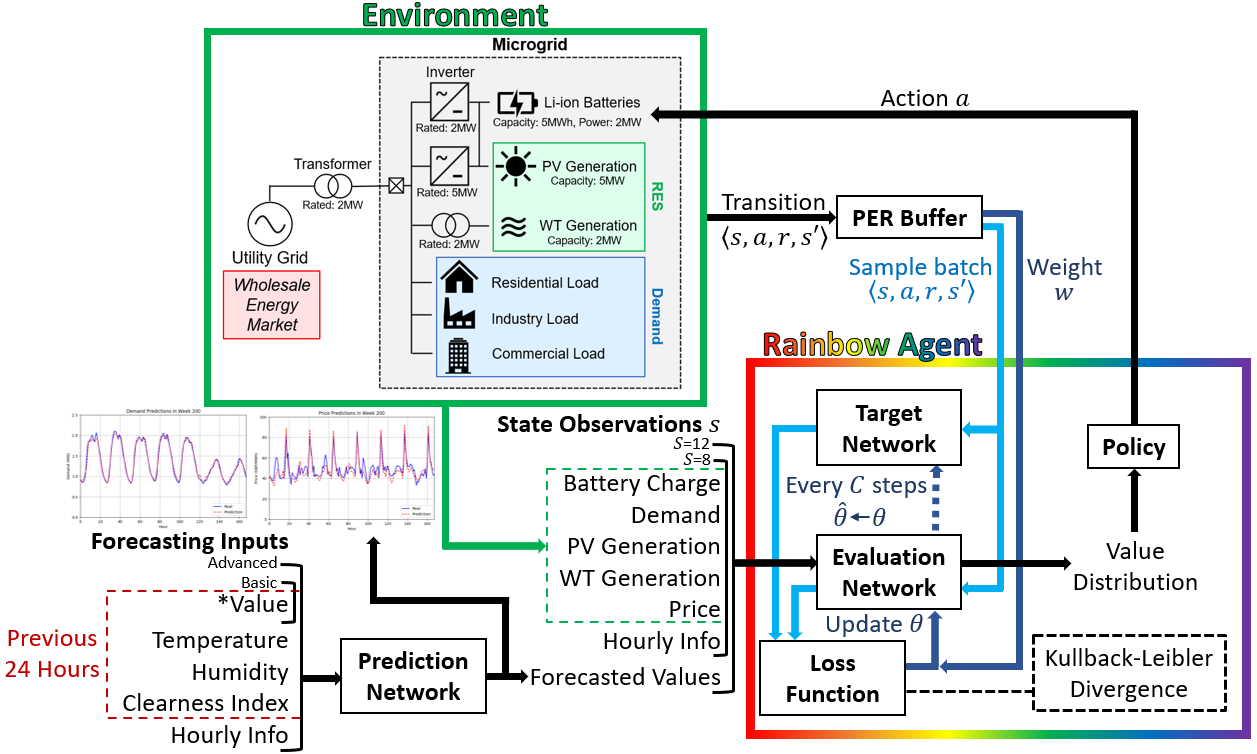}
	\caption{The control process at each step of the environment.} \label{fig:nn}
\end{figure*}

This section outlines the full methodology and setup used in the case study simulations. Figure \ref{fig:nn} shows the process at each time-step. The prediction ANNs provide forecasted values for the state observations which are then passed to the agent to output the action following the learnt policy. The environment then executes that action and the transition is stored in the replay buffer, which is sampled for learning.

\subsection{Microgrid Simulation}
The simulations are run across 4 years of data from 2014 up to 2018 with hourly readings for energy consumption, price, and weather parameters. Each time-step represents one hour with each episode representing one week or 168 time-steps. The simulations are run over 200 episodes with the first 100 are used for training and the second 100 for evaluation. As the environment has no terminal states, the episodes are only relevant for reviewing performance.

Both the forecasting ANNs and algorithms are implemented using \textit{Tensorflow}. For fairness and repeatability, each ANN is set to the same \textit{Tensorflow} and \textit{Numpy} random seed. For the evaluation episodes, the $\epsilon$ value of $\epsilon$-greedy policy methods was set to 0 and the noise was removed from agents with noisy layers. However, the agents will continue to learn during the evaluation steps so are always learning online.

Each method is tested under four different scenarios with different sizes of state and action-space: with and without the forecast predictions as well as with a smaller and larger action-space. Therefore, the agents are tested on how they effectively utilise the forecasted values as well as the resolution of control using a set of discrete actions.

A notable challenge in this case study is the limitation of data. RL agents are often tested using the Arcade Learning Environment \cite{bellemare_arcade_2013} or similar simulated game environments where agents are able to train indefinitely. Due to the limited amount of data available in this case study, it is important that the algorithms are able to train quickly and efficiently.

\subsection{Hyperparamter Testing}
\begin{table}[!b]
\caption{Agent Hyperparameters}
\centering
\begin{tabular}{ll}
\hline
Parameter & Value \\
\hline
ANN Learning Rate & 0.001 \\
Discount Factor $\gamma$ & 0.99 \\
Batch Size & 64 \\
Epsilon $\epsilon$ & 1 $\rightarrow$ 0.01 \\
PER Priority Scale $\alpha$ & 0.7 \\
PER Importance Scale $\beta$ & 0.5 \\
MS-DQN Backup Steps & 2 \\
NN-DQN Initial Noise $\sigma_0$ & 0.5 \\
C51 Atoms & 51 \\
C51 Boundaries $v_\text{min}$, $v_\text{max}$ & -10, 10 \\
DDPG Noise & 0, 0.1 \\
\hline
\end{tabular}
\label{table:hyper}
\end{table}
Different hyperparameters were tested in preliminary simualtions with the final key values shown in  Table \ref{table:hyper}. The PER uses a relitavely high priority scale and importance value with a low offset, all of which promote aggressive sampling of the more important transitions \cite{schaul_prioritized_2016}. Various MS-DQN back-up sizes were tested but performance decreased the larger the step size so Rainbow only uses the standard 1-step backup. Gradient clipping has been suggested to improve the performance of D3QN \cite{wang_dueling_2016} and reward clipping for C51 \cite{bellemare_distributional_2017}. However, these either had no effect or even lowed performance in those preliminary studies as the rewards are typically small between -1 and 1 anyway.

\subsection{Forecasting}
\begin{figure*}[!t]
\centering
	\begin{subfigure}{.48\textwidth}
		\centering
		\includegraphics[width=.9\textwidth]{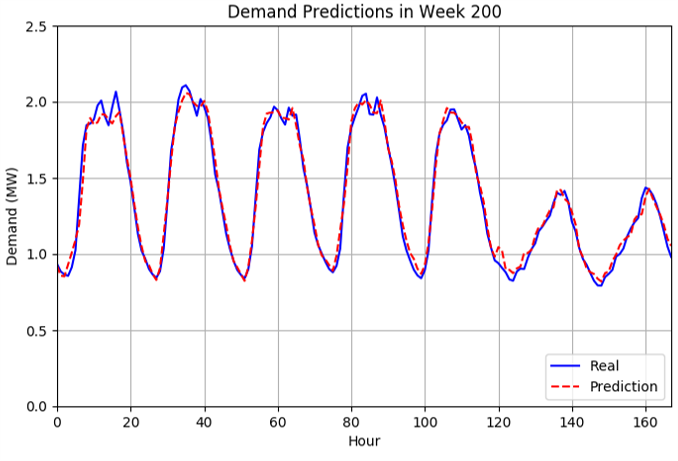}
		\caption{Demand real and predicted values across one week.}
		\label{fig:demand}
	\end{subfigure}
	\begin{subfigure}{.48\textwidth}
		\centering
		\includegraphics[width=.9\textwidth]{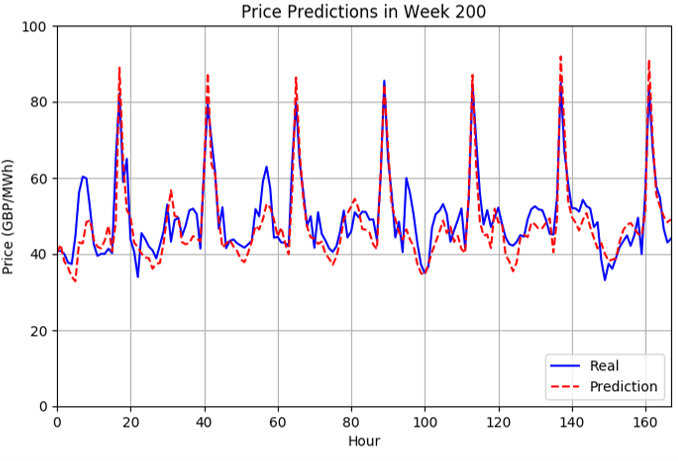}
		\caption{Price real and predicted values across one week.}
		\label{fig:price}
	\end{subfigure}
	\caption{The prediction ANN forecasts for Week 200.}
\end{figure*}

The scenarios with the large state-space use forecasted values calculated using a regression ANN each for demand, price, PV generation, and WT generation. The RES forecasts use basic ANNs provided with the previous 24 hours worth of the values as input. The demand and price prediction ANNs are more advanced and use several weather parameters: dry bulb temperature, air humidity, and sky clearness index. Using the amount of rainfall was also tested but was found to increase the errors due to the high amount of zero-data. The first 100 weeks of the data are used for training and the second 100 for evaluation so that it coincides with the training and evaluating periods of the simulations themselves.

Several different architectures of ANN were tested for the prediction networks including convolutional and recurrent long short-term memory layers for each of the input parameters. However, it was found that a basic ANN had comparable mean absolute percentage errors (MAPE) to the other architectures but often with a significantly faster training time. The MAPE for the used demand predictions is 5.01\% and price predictions is 11.62\%. An example of the values for the demand and price forecasts for the final week given in Figure \ref{fig:demand} and \ref{fig:price} respectively.

\subsection{Benchmarks}
Two benchmarks are used to assess the performance of the DQN agents: an advanced actor-critic method and a linear programming model. Both methods have the advantage of continuous control rather than the discrete set of actions for DQN.

\subsubsection{Deep Deterministic Policy Gradients (DDPG)}
The first benchmark used is DDPG \cite{lillicrap_continuous_2016}, an actor-critic algorithm with many similarities to DQN. The key difference is as well as having a the critic network that estimates the action-value, an actor network is used to calculate a deterministic action with a noise process for exploration.

The main benefit of DDPG over DQN is the continuous action-space meaning it can charge or discharge by any valid amount, whereas DQN must select from one of the discrete actions. However, actor-critic methods typically have much more unstable learning properties than value-function methods. An example of its use for DR is Zhang et al. \cite{zhang_data-driven_2021} as a benchmark for control in a hybrid energy microgrid to compare against a different actor-critic method.

\subsubsection{Linear Programming Model}
The second benchmark is a linear programming model, similar to that used by Cao et al. \cite{cao_deep_2020}. As there are no effective methods for optimally solving general nonlienar programming problems \cite{boyd_convex_2004}, the model uses a simplified environment with constant efficiencies for the ESS, transformer, and inverter. The battery charging and discharging efficiency is generalised to $\eta^{\text{ESS}}=0.95$ with the transformer and inverter efficiency combined and generalised to $\eta^{\text{Grid}}=0.92$.

As the efficiencies are applied differently if the battery is charging or discharging, the power $x_t$ is separated into the charging power $x^{\text{ch}}_t$ and discharging power $x^{\text{dis}}_t$. The model is bound by the following constraints:
\begin{equation}
\text{Constraints}:
\begin{cases}
	0 \leqslant x^{\text{ch}}_t \leqslant \text{X}^{\text{max}} \\
	0 \leqslant x^{\text{dis}}_t \leqslant \text{X}^{\text{max}} \\
	0 \leqslant c_t \leqslant \text{C}_{\text{max}} \\
	c_0 = 0 \\
	c_t = c_{t-1} \eta^{\text{SDC}} + x^{\text{ch}}_t  \eta^{\text{ESS}} - x^{\text{dis}}_t
\end{cases}
\end{equation}

The model performs energy arbitrage using the predicted value of the wholesale energy price at the next step $ \text{P}^{\text{pred}}_{t+1}$. The objective of the solver is to maximise the revenue over the 100 evaluation weeks:
\begin{equation}
\text{Obj} = \max_x \sum\limits_{i=0} \text{P}^{\text{pred}}_i \left( x^{\text{dis}}_i \eta^{\text{Grid}} \eta^{\text{ESS}} - \frac{x^{\text{ch}}_i}{ \eta^{\text{Grid}}} \right)
\end{equation}

The sequence of $x$ values the produced by the model are saved and input into the complete environment for comparison against the agents.
\section{Results and Discussion}
\label{sec:results}
\begin{table*}[!b]
\caption{ESS Savings Results (1000 GBP) and percentage difference against DQN}
\centering
\begin{tabular}{lllllllllll}
\hline
\multirow{2}{*}{Algorithm} & & \multicolumn{4}{c}{Basic ($\mathcal{S}$=8)} & & \multicolumn{4}{c}{Forecasting ($\mathcal{S}$=12)} \\ \cline{3-6} \cline{8-11}
 & & $\mathcal{A}$=5 & vs DQN & $\mathcal{A}$=9 & vs DQN & & $\mathcal{A}$=5 & vs DQN & $\mathcal{A}$=9 & vs DQN \\
\hline
DQN & & 75.00 & - & 68.58 & - & & 78.91 & - & 76.1 & - \\
DDQN & & 71.46 & -4.72\% & 70.88 & 3.35\% & & 81.02 & 2.67\% & 77.97 & 2.46\% \\
D3QN & & 75.11 & 0.15\% & 73.36 & 6.97\% & & 80.69 & 2.26\% & 79.83 & 4.90\% \\
PER-DQN & & 76.69 & 2.25\% & 72.29 & 5.41\% & & 81.55 & 3.35\% & 77.76 & 2.18\% \\
MS-DQN & & 70.55 & -5.93\% & 64.95 & -5.29\% & & 72.56 & -8.05\% & 76.79 & 0.91\% \\
NN-DQN & & 76.47 & 1.96\% & 73.26 & 6.82\% & & 79.65 & 0.94\% & 75.98 & -0.16\% \\
C51 & & 79.25 & 5.67\% & \textbf{81.56} & \textbf{18.93\%} & & 85.28 & 8.07\% & 83.30 & 9.46\% \\
Rainbow & & 79.46 & 5.95\% & 80.00 & 16.65\% & & \textbf{89.33} & \textbf{13.20\%} & \textbf{91.49} & \textbf{20.22\%} \\
\hline
DDPG & & 77.14 & 2.85\% & 77.14 & 12.48\% & & 82.93 & 5.09\% & 82.93 & 8.98\% \\
Model & & \textbf{79.54} & \textbf{6.05\%} & 79.54 & 15.98\% & & 79.54 & 0.80\% & 79.54 & 4.52\% \\
\hline
\end{tabular}
\label{table:results}
\end{table*}

\begin{figure*}[!t] 
	\centering
	\includegraphics[width=.9\textwidth]{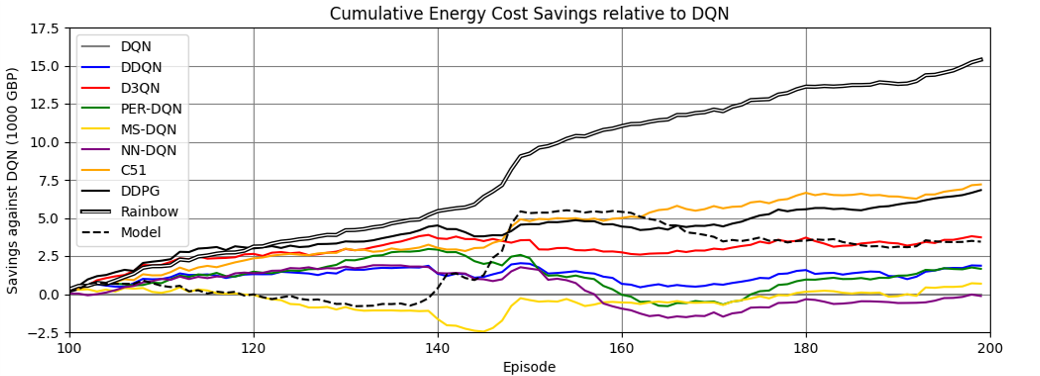}
	\caption{Cumulative energy savings as a percentage relative to DQN across the evaluation episodes.} \label{fig:savingsdqn}
\end{figure*}

\begin{figure*}[!t] 
	\centering
	\includegraphics[width=.9\textwidth]{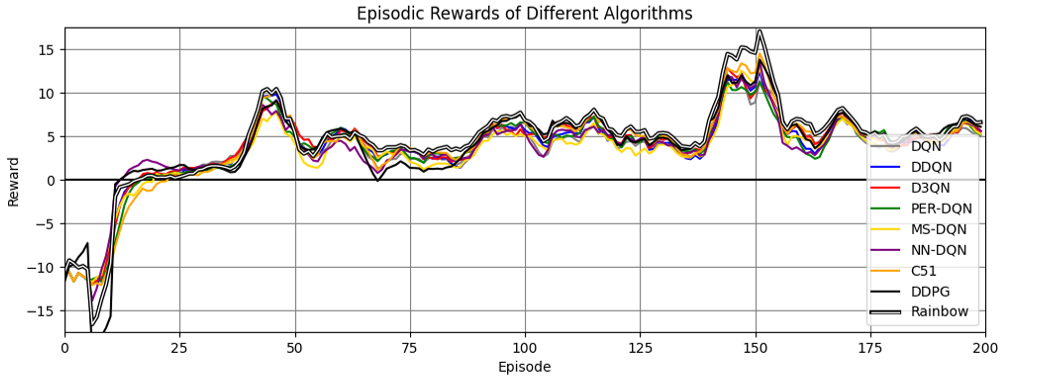}
	\caption{Smoothed episodic rewards (over 5 episodes) of the DRL algorithms across all episodes.} \label{fig:rewards}
\end{figure*}

In these simulation, the DQN algorithms introduced in Section \ref{sec:RL} are used to reduce the energy costs of the microgrid environment outlined in Section \ref{sec:SEN}. The results for four different scenarios using different state and action-spaces can be found in \ref{table:results}. The energy cost savings across the 100 evaluation weeks relative to DQN can be found in Figure \ref{fig:savingsdqn}. The episodic rewards of the different algorithms across all 200 weeks can be found in Figure \ref{fig:rewards}. 

This section will analyse the results in the context of the different algorithms used, how they utilise the large state and action-spaces, and finally looking into the behaviour of the distributional agents.

\subsection{Algorithms}
In the most basic simulation, DQN with the smaller action-space performs largely the same as most of the other DQN variants but performs noticeably worse when either the state or action-space size is increased. It also is never able to achieve a greater saving than the linear programming model. This suggests it is sufficient for simple environments but alternatives should be used for more complex problems.

D3QN and PER-DQN perform better than DQN in all scenarios meaning there is sufficient merit in using the advantage function and a PER respectively. NN-DQN performs better than DQN in the basic scenarios but performs largely the same in the forecasting scenarios. Therefore, an $\epsilon$-greedy policy is sufficient for this application but NN-DQN may complement a PER as the exploration method is tied directly to the learning. However, these methods are only able to outperform the linear programming model in one scenario and cannot outperform DDPG in any. MS-DQN performs worse than all the other methods almost every time suggesting it is inappropriate for this application because the agent cannot see the immediate outcome of its action.

The greatest improvements come from C51 and Rainbow. Both are able to outperform DDPG in every scenario and are only bested by the model solution in the basic environment with the smaller action-space. Rainbow is able to achieve a saving of £91.49k in the forecasting scenario with larger action-space, a 20.22\% increase over DQN’s score. This suggests the various subcomponents of the Rainbow algorithm all complement each other well as the result is much greater than the model, DDPG, and C51.

The distributional approach appears to be the most significant advancement to DQN, as was also found by Hessel et al. \cite{hessel_rainbow:_2018} for Atari games. For this application, we suggest the distributional approach is better able to interpret the punishments for exceeding the ESS’s capacity limits which would affect the expectation for a direct action-value calculation, as well as learning how to navigate the nonlinear efficiency profiles of the environment’s components.

\subsection{Action-Space and Forecasting}
\begin{figure*}[!t] 
	\centering
	\includegraphics[width=.90\textwidth]{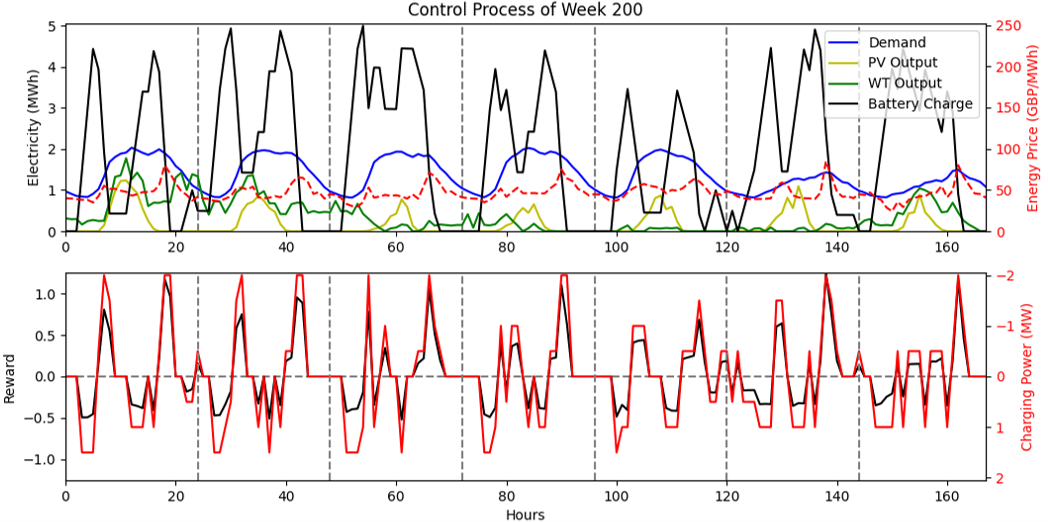}
	\caption{Control process at each step of Episode 200 using Rainbow DQN.} \label{fig:control}
\end{figure*}

Most of the DQN variants all had reduced performance when the action-space $\mathcal{A}$ was increased from 5 to 9, meaning the agent could not explore enough state-action pairs as effectively when there are more actions to select from. The two notable exceptions to this are C51 and Rainbow where their performance either stays largely the same or even slightly increased. This is particularly interesting as it suggests the distributional approach to value calculation is more effective in larger action-spaces and can give finer control of the environment than regular DQN. Despite the model and DDPG have a continuous action-space, the superior learning properties of Rainbow and C51 proved more valuable.

All of the agents performed better when provided with future forecasted values but the most notable improvement was in the Rainbow agent. The various subcomponents were able to combine well and proved effective together in interpreting and learning from the forecasted values with Rainbow performing significantly better than all other DQN variants and the two benchmarks. An example of the control process of the Rainbow agent for the final week of the forecasted state-space and larger action-space can be found in Figure \ref{fig:control}.

\subsection{Value Distribution}
\begin{figure*}[!t]
\centering
	\begin{subfigure}{.45\textwidth}
		\centering
		\includegraphics[width=\textwidth]{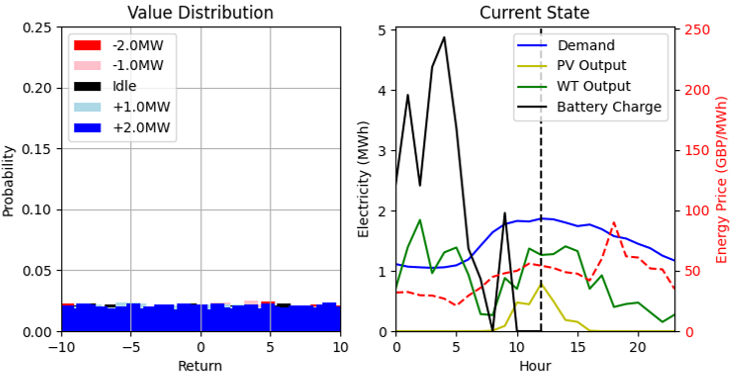}
		\caption{Episode 1, Day 2.}
		\label{fig:cat1}
	\end{subfigure}
	\begin{subfigure}{.45\textwidth}
		\centering
		\includegraphics[width=\textwidth]{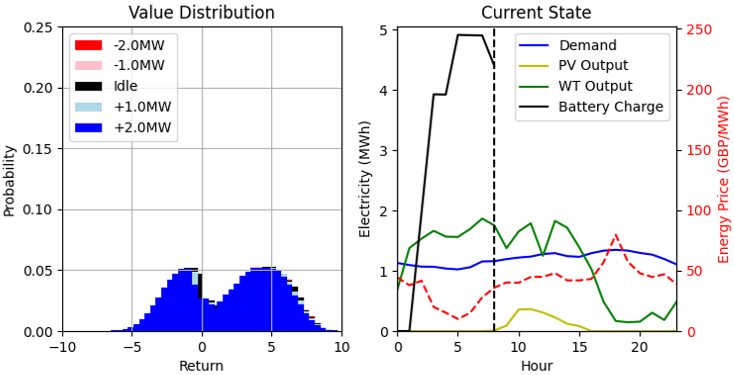}
		\caption{Episode 50, Day 6.}
		\label{fig:cat2}
	\end{subfigure}
	\begin{subfigure}{.45\textwidth}
		\centering
		\includegraphics[width=\textwidth]{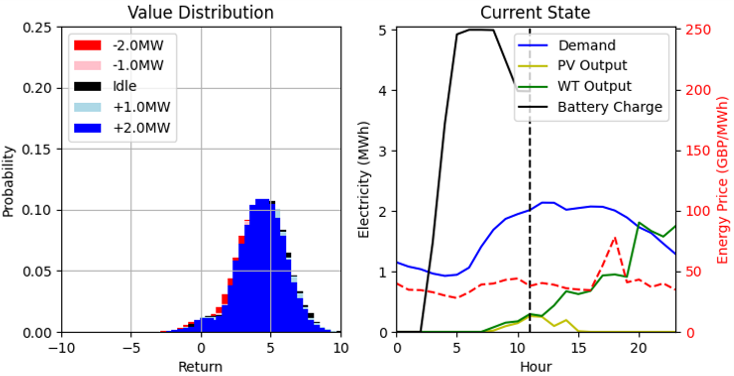}
		\caption{Episode 100, Day 4.}
		\label{fig:cat3}
	\end{subfigure}
	\begin{subfigure}{.45\textwidth}
		\centering
		\includegraphics[width=\textwidth]{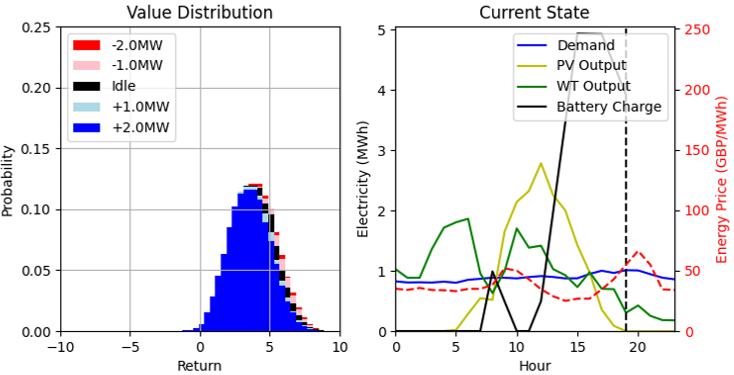}
		\caption{Episode 170, Day 7.}
		\label{fig:cat4}
	\end{subfigure}
	\begin{subfigure}{.45\textwidth}
		\centering
		\includegraphics[width=\textwidth]{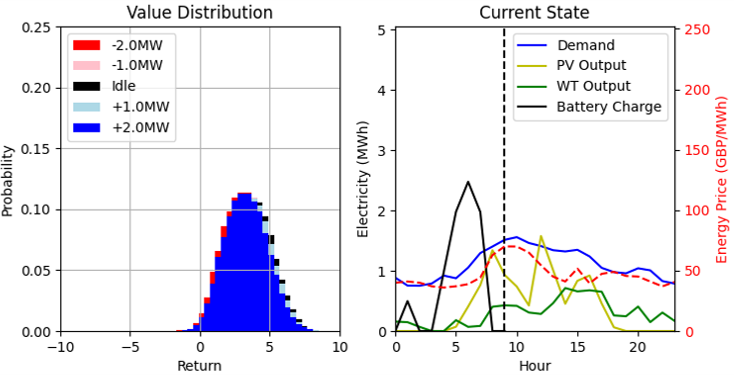}
		\caption{Episode 190, Day 2.}
		\label{fig:cat5}
	\end{subfigure}
	\begin{subfigure}{.45\textwidth}
		\centering
		\includegraphics[width=\textwidth]{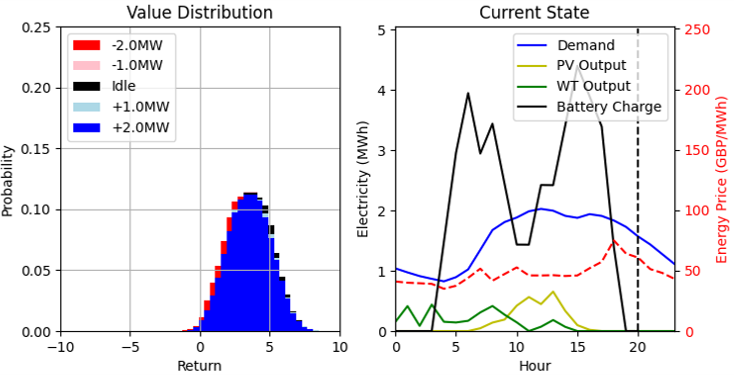}
		\caption{Episode 200, Day 4.}
		\label{fig:cat6}
	\end{subfigure}
	\caption{Value distributions across different episodes, with the current step marked by the black dotted line.}
	\label{fig:dist}
\end{figure*}

Figure \ref{fig:dist} shows how the Rainbow agent learns the value distribution over time at different steps. In each of the subfigures, the plot on the left shows the probability of getting a particular return while the plot on the right shows the state of the environment at that step. For context, the value distribution of less valuable actions appear to the left with more valuable actions to the right. Despite the figures being taken in the larger action-space scenario, the distributions of only five actions have been shown in the interest of legibility.

At episode 1 (\ref{fig:cat1}) the distribution is random as the agent has not yet begun learning. By episode 50 (\ref{fig:cat2}), the agent has begun to understand the environment but there is still a probability of taking a negative action; such as charging or discharging when the ESS is full or empty respectively. By episode 100 (\ref{fig:cat3}), the agent has learnt that it can almost always take a positive action in any state.

Figure \ref{fig:cat4} shows an example of a step where the ESS has a lot of charge before a period of higher energy prices with the higher return of the discharging actions showing the agent has learnt to discharge in that event. Figure \ref{fig:cat5} show a case where the ESS is empty but prices are relatively high so remaining idle is the most valuable action. Figure \ref{fig:cat6} shows a case at the end of the day as the prices begin to drop where the agent has learnt its most valuable to either remain idle or begin charging more slowly.

These findings are important as it shows the benefits of using the distributional RL approach. Whereas most methods learn the expectation of the action-value, Rainbow and C51 can better understand non-deterministic reward functions and under what conditions those irregularities occur; such as the punishment received for exceeding the ESS's boundaries in this case study. The distributional approach also provides a means to visualise how the agent is learning, which is valuable for contextualising and interpreting the behaviour of RL agents.

\subsection{Improvements and Further Work}
One improvement that could be made to this case study is more rigorous testing of the forecasting methods with the breadth of historical weather data available. However, the accuracy of the prediction ANNs were more than sufficient in this case study as a supplement to the RL agents.

In the future, this case study will be built upon using a multi-agent system. Here, different agents will be able to control multiple different types of ESS collaboratively alongside forms of distributed generation. This may also include competitive agents with selfish interests for other stakeholders; such as home owners with their own smaller ESSs or for the charging of electric vehicles.
\section{Conclusion}
\label{sec:conclusion}
In this paper, the deep RL algorithm Rainbow DQN was used to operate an ESS in a microgrid with its own demand, renewable energy, and dynamic energy prices. The various subcomponents of Rainbow were tested and found that the value distribution approaches - C51 and Rainbow - outperformed the other algorithms by a notable margin. Basic DQN was sufficient in the simpler scenarios but struggled in the more complex cases. The components of Rainbow were able to effectively compliment each other and allowed the agent to effectively use and learn from the future forecasting values, outperforming the other DQN variants and the benchmarks showing the superior learning capabilities of Rainbow. By plotting the value distribution, it was also found to be an effective means to contextualise and understand the agent's behaviour.

\section*{Acknowledgements}
This work was supported by the Smart Energy Network Demonstrator project (grant ref. 32R16P00706) funded by ERDF and BEIS. This work is also supported by the EPSRC EnergyREV project (EP/S031863/1) and the Royal Society Research Grant (RGS/R1/191395).

\bibliographystyle{elsarticle-num}
\bibliography{ESS_Project}

\end{document}